# Lane detection with Position Embedding

Jun Xie, Jiacheng Han, Dezhen Qi, Feng Chen, Kaer Huang, Jianwei Shuai

PCIE Lab, Lenovo Research

## ABSTRACT

Recently, lane detection has made great progress in autonomous driving. RESA (REcurrent Feature-Shift Aggregator) is based on image segmentation. It presents a novel module to enrich lane feature after preliminary feature extraction with an ordinary CNN. For Tusimple dataset, there is not too complicated scene and lane has more prominent spatial features. On the basis of RESA, we introduce the method of position embedding to enhance the spatial features. The experimental results show that this method has achieved the best accuracy 96.93% on Tusimple dataset.

## 1. INTRODUCTION

Autonomous driving technology is the product of the combination of automotive industry technology and artificial intelligence technology, which includes three main systems: perception system, decision-making system, and control system. The perception system obtains information about the road environment around the vehicle through various sensors on the vehicle body. The existing intelligent driving is usually equipped with LDWS (Lane Departure Warning System) and LKA (Lane Keeping Aid). The core of these systems is lane detection algorithm.

As a basic task of road environment perception, lane detection plays a key role in the perception system of autonomous driving. At present, the main difficulty of the lane detection is that its real-time and accuracy cannot be balanced, especially in some complex environments. Most of the traditional detection algorithms [1,2,3] affected by environmental changes, and problems such as false detection and missed detection are prone to occur. In recent years, deep learning technology has made significant achievements in many fields and provides new ideas for the research of lane detection [4, 5, 6, 7, 8]. In deep learning, convolutional neural network (CNN) is mainly used for image feature learning.

The main problems faced by lane detection are: (1) lane features are thin and long in the picture; (2) sectionalized and discontinuous; (3) blocked by vehicles severe occlusion; (4) lane is partially damaged and ambiguous; (5) disturbed by duzzle light or crowded environment.

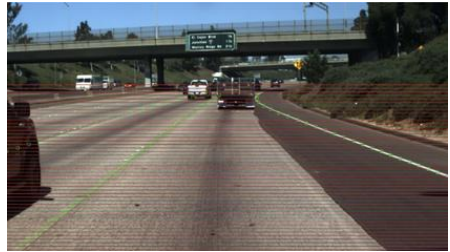

Figure 1. Example for Tusimple dataset

There are currently two general datasets, Tusimple [9] and CULane [10]. The Tusimple scene come from highway (shown in Figure 1). The characteristics are as follows: (1) Good and moderate weather conditions; (2) Different time periods during the day; (3) 2 lanes / 3 lanes / 4 lanes / or more.

The CULane data set includes 9 scenes (Figure 2). The scenes are complex. Crowded, Nights, and No line are the harder scenes to detect lane in this data set. We focus on the Tusimple dataset.

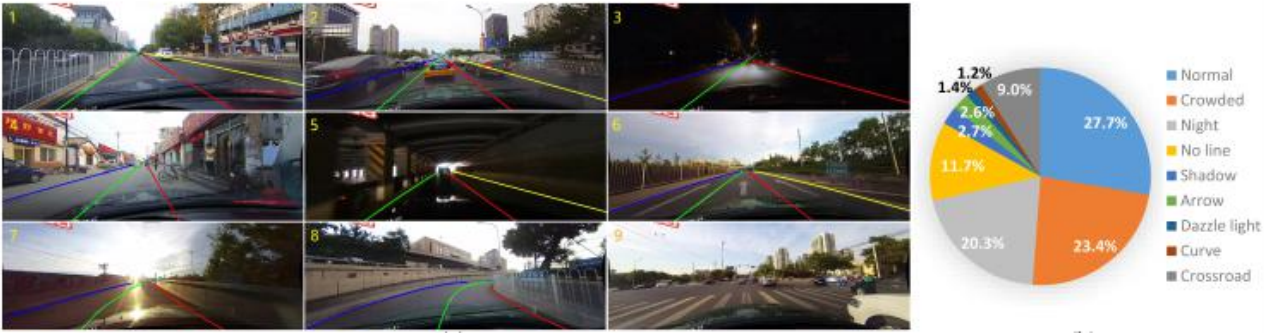

Figure 2. Dataset examples for different scenarios; Proportion of each scenario

## 2. RELATED WORK

There are currently several types of deep learning methods applied to lane detection, Semantic/instance segmentation-based [4, 10, 11, 12], key point detection [5, 13], gridding [6, 14, 15], Ploynominal [7, 16], Anchor-based [8], etc.

Focus on Local [13] is inspired by the human body key point detection algorithm, directly detects the key points of the lane line, and uses local geometry construction to correct the position of the key points. It achieves the best state-of-the-art level with accuracy 96.92% on Tusimple.

PINet [5] uses a key point detection method based on Stacked Hourglass Network. To improve the accuracy of predicted points by predicting the offset in the x and y directions, using L2 feature loss to cluster different lane lines. It is obtaining 96.75% Accuracy in the Tusimple dataset.

CondLaneNet's backbone [6] is based on the original conditional instance segmentation strategy, CondLaneNet include Proposal head and Conditional shape head, Proposal head produces lane segmentation instance, Conditional shape head is responsible for point position in lane. CondLaneNet's row-wise network is more fit for CULane's various scenario. It gets F1 score: 85.09, for simple Scenario in Tusimple, Acc:96.54%. CondLaneNet is not best performance compare with FOLOLane or RESA for lack of global spatial information on Tusimple. Pictures in Tusimple data have more structured information than picture in CULane. FOLOLane or RESA is more fit for such scenario.

RESA, SCNN uses the method of instance segmentation. When people recognize lane lines, they can infer the entire lane line through partial clues and spatial correlation. Therefore, RESA and SCNN all use the spatial information aggregation. The spatial information is aggregated to enhance the characteristics of lane lines. In addition, Resa uses Bilateral Up-Sampling Decoder, which is divided into coarse grained branch and fine detailed branch, this method guarantees better detailed information after up-sampling. The results of the paper show that the method of instance segmentation achieves the second Accuracy, Acc:96.82, in a single scene like Tusimple.

## 3. METHOD

The Tusimple data set itself has a single scene, with less lane line occlusion and no hard light interference. RESA has done feature aggregation on Tusimple. Through the visualization, after the RESA operator, there are already obvious line space features. With further observation, the Tusimple data set also has strong spatial characteristics, including (1) the lane lines are relatively "parallel" (2) the starting points of several lane lines are all below the picture (3) the convergence point of multiple lane lines in the distance (Vanishing Point). CNN itself can learn location features [17, 18 19]. The classification and detection mainly use the equivariance to translation of CNN, which can be achieved by using relative position coding instead of absolute position coding [20, 21]. Inspired by the role of position embedding in natural scene text recognition [22, 23], we guess that lane line scenes has similar characteristics with natural scene text recognition lines, so Position Embedding is introduced to describe spatial information. We have verified Four methods, Sinusoidal Position Encoding.

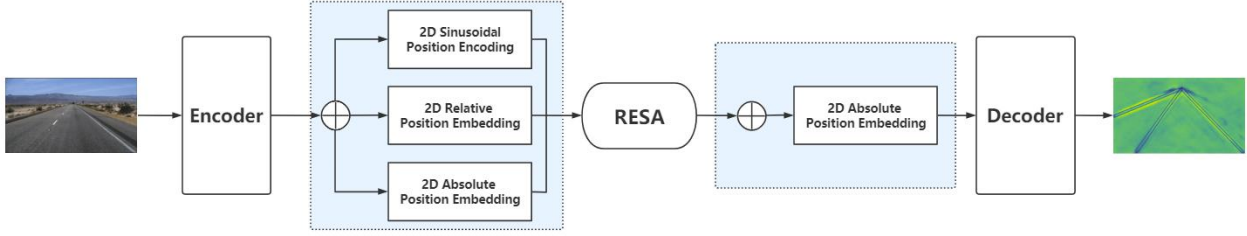

Figure 3. Architecture Design

### 3.1 Sinusoidal Position Encoding

[22, 23]believe that the self-attention blocks are similar to a fully connect layer, agnostic to spatial arrangements of its input. Input the feature map to the self-attention block, which can provide long-term dependencies that cannot be captured by the normal convolution. In our task, lane line has a similar spatial feature distribution to the text in the natural scene. On the other hand, Self-Attention does not provide absolute position information, so it cannot distinguish the positional relationship well between lane lines. And the position information plays an important role in identifying the slender shape of the lane line. Therefore, we use a 2D Absolute Sinusoidal Position Encoding. The network architecture is shown in Figure 3.

Sinusoidal Position Encoding was first proposed by avaswani [24]. In general, for each point $x_i$ of the feature map $x = (x_1, x_2, ..., x_n)$, add a positional encoding $p = (p_1, p_2, ..., p_n)$ as

$$x_i = x_i + p_i \qquad (1)$$

where the positional encoding $p_i, x_i$.

The simplest example of positional encoding is an ordered list of values, between 0 and 1. Here choose the fixed encoding by sine and cosine functions with different frequencies:

$$PE_{(pos,\ 2i)} = \sin\left(pos/10000^{2i/d_{model}}\right) \qquad (2)$$
$$PE_{(pos,\ 2i+1)} = \cos\left(pos/10000^{2i/d_{model}}\right) \qquad (3)$$

Where $PE_{(pos,2i)}$ and $PE_{(pos,2i+1)}$ respectively are the $2i, 2i+1$ component of the encoding vector at position pos, i is the dimension of the channel. That is, each dimension of the positional encoding corresponds to a sinusoid.

Compared to coordinate encoding, is kind of embedding manifold which is a way to make the position encoding a discretization of something continues. Our main contribution is to introduce absolute position information encoded by Sinusoidal for lane line recognition tasks and capture long-term dependencies through self-attention.

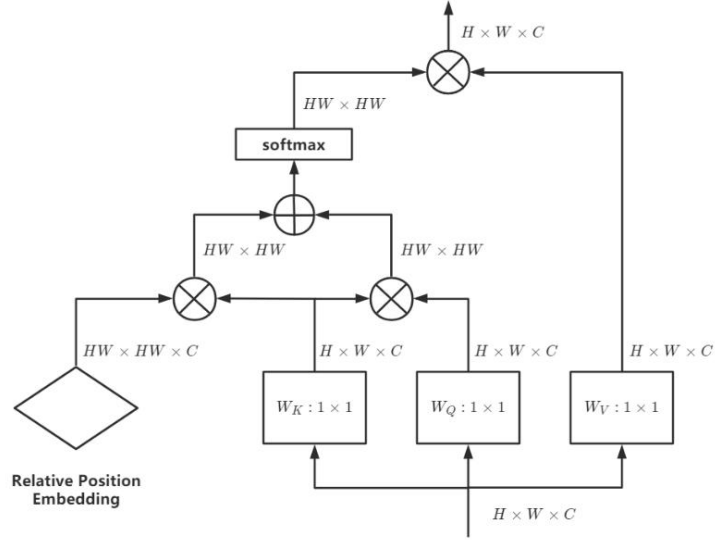

Figure 4. Illustration of self-attention modules with 2D relative position encoding on keys

### 3.2 Relative Position Embedding

Recently, more and more work [20,21, 25, 26] has focused on the position embedding of relative relations, and found that relative position embedding [27] is more suitable for visual tasks. The self-attention block with relative position embedding provides long-term content dependency while also considers the positional relationship between different distance features, therefore, it is effective to associate content information with location information.

Relative position embedding was first proposed by Shaw [27]. By encoding and training the positional relationship between features at different distances enhanced the relative position information and satisfy translation equivariance.

We introduce relative position embedding in our work. For feature map that have been down-sampled by the encoder:

$$x = \begin{bmatrix} x_{11} & \cdots & x_{1w} \\ \vdots & \ddots & \vdots \\ x_{h1} & \cdots & x_{hw} \end{bmatrix}_{h \times w}$$

expand it to $x = (x_{11}, x_{12}, ..., x_{1w}, x_{21}, ..., x_{2w}, ..., x_{h1}, ..., x_{hw})_{h \times w}$

Self-attention block receives x and outputs $z = (z_1, z_2, ..., z_{h \times w})_{h \times w}$. Each output element $z_i$ is computed as weighted sum of a linearly transformed input elements:

$$z_i = \sum_{j=1}^{n} \alpha_{ij}(x_j W^V) \qquad (4)$$

Each weight coefficient $\alpha_{ij}$ is computed using a SoftMax:

$$\alpha_{ij} = \frac{\exp(e_{ij})}{\sum_{k=1}^{n} \exp(e_{ik})} \qquad (5)$$

Where $e_{ij}$ is calculated using a scaled dot-product attention with relative position embedding:

$$e_{ij} = \frac{(x_i W^Q)(x_j W^K)^T + (x_i W^Q) r_{ij}^T}{\sqrt{d_z}} \qquad (6)$$

Here, the projections $W^Q, W^K, W^V$ are parameter matrices, which are unique per layer. $r_{ij}^T$ means the relative position information matrix.

Compared with classification tasks that require translation invariance, lane line recognition is a semantic segmentation task. We believe that it needs translation and other changes. The relative position embedding just satisfies this property, so the relative position embedding can be used to introduce position information while enhance the translation and other degeneration.

### 3.3 Absolute and Relative Position Embedding

Considering that lane detection is different from general semantic segmentation tasks, the position of lane lines in the image is relatively fixed and spans the entire image. For segmentation task in coco dataset, A cat can appear anywhere in one picture, so relative position embedding is fit for capturing the feature of the cat. But for lane line in Tusimple dataset, lane line is usually full of picture and spread form bottom of picture to top part. It has more "fixed" spatial feature than picture in coco dataset. Therefore, we believe that absolute position information plays a vital role for our task. Based on this assumption, we added absolute position embedding introducing relative position embedding.

### 3.4 Absolute Position Embedding

Both RESA and Self-attention focus on solving the drawback of CNN: a large receptive field is needed to track the long-range dependencies within the image. Increasing the size of the convolution kernel increases the network's representation ability while losing the computational and statistical efficiency obtained by using a local convolution structure. That is, the problem is that CNN cannot obtain remote information efficiently.

Self-attention calculates the similarity between each point in the feature map and all other points, and expresses this similarity with scaled dot-product, and then uses the obtained similarity to add the information of all points to each point. From a certain perspective, Self-attention is a large-scale convolution kernel, which extracts global features, and performs feature aggregation under the guidance of the Self-attention mechanism. There are two other method of feature aggregation. Pan [10] propose Spatial CNN (SCNN), enabling message passing between pixels across rows and columns in a layer. Fang [12] develop a REcurrent Feature-Shift Aggregator (RESA) to do global feature aggregation.

Lane detection is an extremely special semantic segmentation problem, which is inherently slender and often interrupted by environmental influences. And due to the thin and long property of lanes, the number of annotated lane pixels is far fewer than background pixels. In this case, do attention to each point is not so suitable. However, RESA slices the feature map horizontally and vertically, transmits information in the four directions of rows and columns, and aggregates global information. This approach can greatly increase the dependence on the surrounding information. That is to say, when distinguishing lane line points, visual clues such as other lane lines and the shape of the road can be used to help the detection. And because of its relatively high computational efficiency, multiple iterations can be performed, so that high-level semantic information can be obtained, which is a great help for inference and judgment.

Therefore, we consider removing Self-attention block here and directly enhancing the position information on the down-sampled feature map. And considering that after RESA, the spatial structure information of the overall feature map has been greatly enhanced (it can also be seen from the Figure 5), so we choose to add position information after RESA, which can make the position enhancement play a better role.

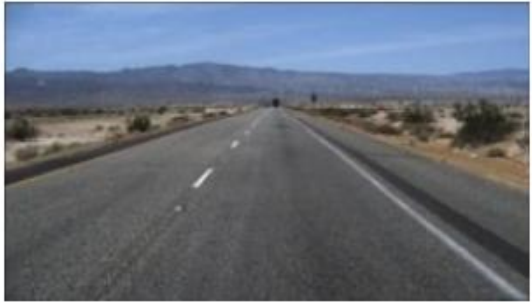
(a) Natural

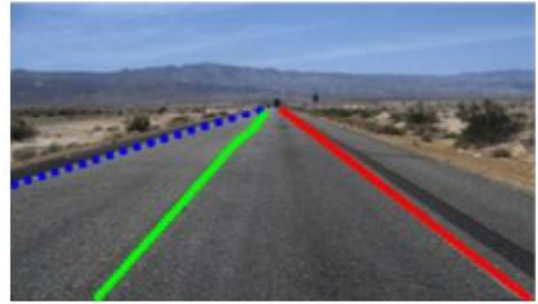
(b) GT

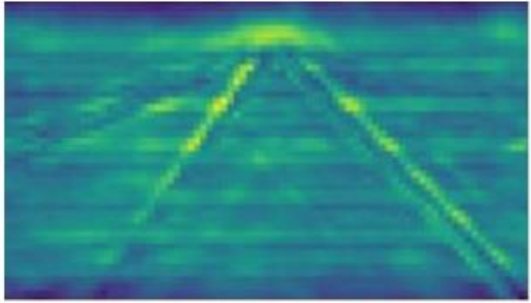
(c) RESA

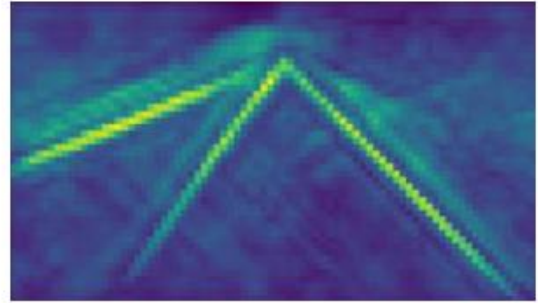
(d) RESA with PE

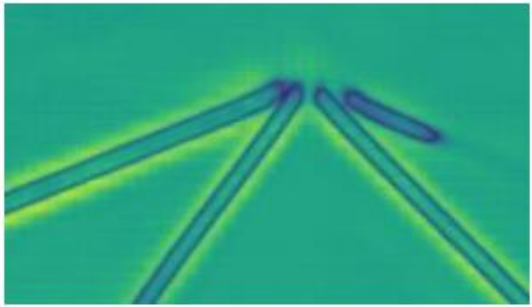
(e) Deocder

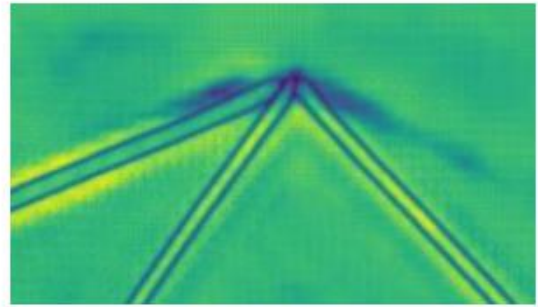
(f) Decoder with PE

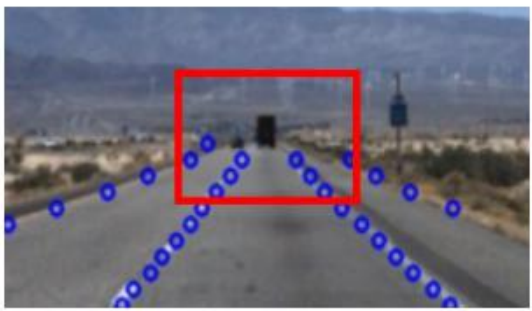
(g) Prediction(without PE)

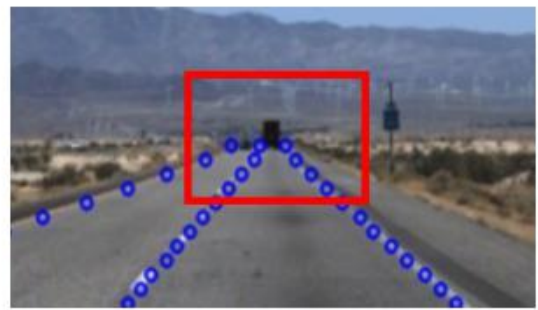
(h) Prediction(with PE)

Figure 5. Visualization

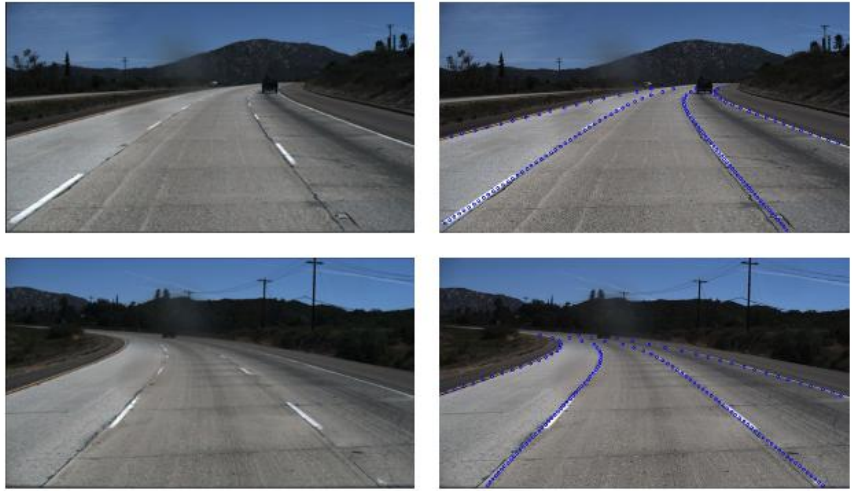

Figure 6. Other different kinds of lane

## 4. VISUALIZATION

Figure 5 shows the visualization of feature maps after RESA blocks. The right image Figure 4 (d) clearly shows that the addition of absolute position embedding greatly enhances the sensitivity of the model to lane lines, while reducing the redundancy of information stacking in the background part introduced by RESA.

Figure 5 (d) clearly shows that the color of the lane lines is brightened, while the background color is darkened. The overall outline of the lane line is clear, and the overall feature map becomes more structured.

Figure 5 (f) shows the feature map after down-sampling. It can be seen that after adding the absolute position information, the corresponding position of the lane line is obviously prominent, and the outline of the lane line is clearer. And for the wrongly recognized lane line, it is prohibited. [28]show that by training a deep network, the overall representation of shallow features by the network is also improved. Therefore, the location information introduced in this paper also enhances the RESA module and improves the aggregation effect.

At the same time, it can be considered that the position embedding the prior information that the lane lines are merged into a point. Therefore, the prediction of the long-range convergence part in the subsequent lane line point prediction is more detailed and accurate (see Figure 5 (h)).

Figure 6 shows the recognition effect of other different kinds of lane lines, such as curved non-straight lines and dashed lines. The Tusimple dataset does not distinguish between dashed and solid lines, and since our method adopts the idea of segmentation, it can effectively deal with scenes with curved lane lines.

## 5. EXPERIMENT

### 5.1 Dataset and Metrics

To evaluate the performance of our proposed method, we conduct experiments on widely used detection benchmark datasets: Tusimple, which is collected with stable lighting conditions in highways. For Tusimple dataset, the evaluation metric is accuracy. It is defined as follow:

$$accuracy = \frac{\sum_{clip} C_{clip}}{\sum_{clip} S_{clip}} \qquad (7)$$

which $C_{clip}$ is the number of lane points predicted correctly (mismatch distance between prediction and ground truth is within a certain range) and $S_{clip}$ is the total number of ground truth points in each clip.

All models are trained with docker on NVIDIA Tesla V100 GPUs (32G Memory) in Ubuntu. All experiments are implemented with Pytorch1.10.

## 5.2 Main Result

We compare our approach with state-of-the-arts including SCNN [10], PINet [5], RESA [12], FOLOLane [13] on Tusimple dataset, as listed in Table 1( * represents that the data is not provided in the original paper). Our method is based on RESA. It can be seen that compared with RESA, our four method are improved and almost outperforms other methods. And it is worth noting that our absolute position embedding (APE) method and the combined absolute position and relative position (RPE+APE) method outperform the state-of-the-art methods. However, the introduction of relative position information will consume a lot of gpu resources. The gpu occupies 14225MiB, and the introduction of relative position embedding consumes 19135MiB. These observations demonstrate the effectiveness and robustness of our method and validate the effectiveness of PE on the lane detection task.

Meanwhile, our method can achieve 35fps, that is, adding our location information does not affect the efficiency. Our method can be applied to real-time applications.

We train RESA on CULane dataset and get F1 score 74.65. We get F1 score 75.27 with position embedding module. Meanwhile, our method can achieve 40fps.

Table 1. Comparison with state-of-the-art results on Tusimple dataset.

| Network | Accuracy | FPS |
|---|---|---|
| SCNN | 96.53 | * |
| PINet | 96.75 | 35 |
| RESA-34 | 96.82 | 35 |
| FOLOLane | 96.92 | * |
| Ours (Sin-PE) | 96.86 | 35 |
| Ours (RPE) | 96.88 | 28 |
| Ours (RPE+APE) | 96.92 | 27 |
| Ours (APE) | 96.93 | 35 |

## 5.3 Ablation Study

In order to verify the effectiveness of our method, we conducted several experiments on the Tusimple dataset for verification (shown in Table 2). The batch size is set 12 and the epoch is 850 for Tusimple. All experiment are test with GeForce RTX 2080 12G GPUs in Ubuntu. We get better accuracy 96.85 than the 96.82 in paper REcurrent Feature-Shift Aggregator because more batchsize and epoch is used. At the same time, adding the PE (APE) module has significantly improved by 0.8.

Table 2. Experiments of the proposed modules on Tusimple dataset(√ representative adoption module).

| RESA | PE (APE) | Accuracy |
|---|---|---|
|  |  | 96.33 |
| √ |  | 96.85 |
| √ | √ | 96.93 |

# 6. CONCLUSION

In this paper, we propose a method for lane line recognition: enhancing location information on the basis of RESA. We believe that absolute position information is very important for this kind of special semantic segmentation problems. We have conducted a large number of experiments to verify our ideas and conducted detailed visual analysis. Through the absolute position embedding, we achieved the first place on the Tusimple dataset.